\documentclass[10pt,twocolumn,letterpaper]{article}

\usepackage{cvpr}
\usepackage{times}
\usepackage{epsfig}
\usepackage{graphicx}
\usepackage{amsmath}
\usepackage{amssymb}
\usepackage{subcaption}
\usepackage{makecell}
\usepackage{amsmath,amssymb}
\usepackage[pagebackref=true,breaklinks=true,letterpaper=true,colorlinks,bookmarks=false]{hyperref}

\cvprfinalcopy % *** Uncomment this line for the final submission

\def\cvprPaperID{****} % *** Enter the CVPR Paper ID here

\ifcvprfinal\fi
\begin{document}

%%%%%%%%% TITLE
\title{Im2Mesh GAN: Accurate 3D Hand Mesh Recovery from a Single RGB Image}

\author{Akila Pemasiri \hspace{1cm} Kien Nguyen Thanh \hspace{1cm} Sridha Sridharan \hspace{1cm} Clinton Fookes \\
Speech, Audio, Image and Video Technology (SAIVT) Laboratory\\
Queensland University of Technology, Brisbane, Australia\\
{\tt\small \{a.thondilege, k.nguyenthanh, s.sridharan, c.fookes\}@qut.edu.au}
}
\maketitle
%\thispagestyle{empty}

%%%%%%%%% ABSTRACT
\begin{abstract}
\vspace{-10pt}
This work addresses hand mesh recovery from a single RGB image. In contrast to most of the existing approaches where the parametric  hand models are employed as the prior, we show that the hand mesh can be learned directly from the input image. We propose a new type of GAN called Im2Mesh GAN to learn the mesh through end-to-end adversarial training. By interpreting the mesh as a graph, our model is able to capture the topological relationship among the mesh vertices. We also introduce a 3D surface descriptor into the GAN architecture to further capture the 3D features associated. We experiment two approaches where one can reap the benefits of coupled groundtruth data availability of images and the corresponding meshes, while the other combats the more challenging problem of mesh estimations without the corresponding groundtruth. Through extensive evaluations we demonstrate that the proposed method outperforms the state-of-the-art.
\vspace{-12pt}
\end{abstract}

%%%%%%%%% BODY TEXT
\section{Introduction}
\vspace{-5pt}
Compared with existing research on 2D or 3D hand pose estimation from RGB or/and depth image data, hand mesh recovery can provide a more expressive and useful  representation for monocular hand image understanding. The hand mesh recovery from a single RGB image is of particular interest for a wide range of applications in many domains, including augmented reality \cite{lee2009multithreaded,piumsomboon2013user} and human computer interaction \cite{markussen2014vulture,sridhar2015investigating}.

However, hand mesh recovery is an ill-posed problem considering multiple meshes can be inferred from the same RGB image. The popular solution to deal with this ill-posed recovery is using priors. Most of the hand mesh recovery approaches in the literature employ the parametric MANO hand model \cite{MANO:SIGGRAPHASIA:2017} as the hand prior and employ some forms of neural networks to regress the model parameters \cite{3DHandShapePoseWild,HandMeshObjects,HAMR}. However low dimensional nature of the parametric models limits their capability to capture non-linear shapes of hands \cite{ge20193d}. In addition, some approaches rely on the heatmaps of the  keypoint annotations in the early steps of their model \cite{HAMR,ge20193d}. We argue that this is redundant since the 3D keypoints can be learned simultaneously with the mesh and they should be learned simultaneously due to the complementary nature of two tasks.

In this paper, we approach the problem of learning the priors by end-to-end adversarial training. We show that the hand priors can be learned explicitly in the 3D mesh representation and can be encoded in a generative network. We propose a new type of Generative Adversarial Network (GAN) called Im2Mesh to learn the mesh vertices directly from a single RGB input image. Through the competing process of the generator and discriminator, the generator gradually improves to a level where it can generate the mesh directly from a single input image, providing an accurate solution for the hand mesh recovery task. 

Importantly, by interpreting  the mesh as a graph, we can employ recent advances in Graph Neural Networks (GNNs) to support mesh processing in both generator and discriminator networks. GNNs have demonstrated the capability of handling non-Euclidean structured data such as graphs and manifolds \cite{monti2017geometric,wu2019comprehensive}. In contrast to the existing graph-based mesh estimation methods in the literature \cite{ge20193d} which only consider the CNN generated features, we introduce a 3D descriptor that encodes surface level information into the GNNs, allowing them to better exploit the topological relationship among mesh vertices in the graph-structured hand data. This improves the mesh recovery accuracy since the recovery algorithm not only considers the vertex 3D coordinates but also the 3D features associated with the vertices. 

Our main contribution of this paper are summarised as below:\begin{itemize}
    \item We propose a new GAN architecture called Im2Mesh to enable end-to-end learning of the hand mesh directly from a single RGB input image, without requiring any heatmap processing, 3D keypoint annotations or external parametric hand models.  \vspace{-5pt}
    
    \item We model the generator of GAN a graph architecture, allowing it to model the topological relationship among the vertices of the mesh while introducing a 3D descriptor to encode the surface level information into the GNNs, further capturing the 3D features associated with the mesh vertices.  \vspace{-5pt}
    
    \item The proposed approaches not only address the problem of mesh reconstruction for the coupled datasets where one to one mapping  prevails between the images and the groundtruth meshes, but also address the problem of reconstructing mesh for the datasets which do not contain the corresponding groundtruth annotation.  \vspace{-5pt}
    
    \item We do not use the depth images; as such we increase the potential of using our model for the datasets which do not have the corresponding depth images. \vspace{-5pt}

\end{itemize} 
The remainder of this paper is organised as follows. In Section 2 we discuss the related work in the research area and their limitations. Section 3 describes our methodology with subsections outlining each component where we describe the approach for the coupled data as well as for  uncoupled data. In Section 4 we present our experimental results, and we conclude the paper in Section 5 while describing the results of the proposed methodology. \vspace{-6pt}
%--------------------------------------to-----------------------------------

\section{Related Work}
\noindent \textbf{3D hand pose and mesh estimation using parametric models:} Majority of existing 3D hand pose and shape estimation methods  \cite{moon2020i2l,HAMR} are based on MANO  \cite{MANO:SIGGRAPHASIA:2017}    model which is a low dimensional parametric representation of hand mesh. However, there remain few weaknesses in using  such parametric models. Firstly, the model is generated in controlled environments which are different from the images that are encountered in real world \cite{choi2020pose2mesh} thus causing a domain gap. Secondly,  the low dimensional nature of the parametric models limits their capability to capture non-linear shapes of hands \cite{ge20193d}. Thirdly, to create a parametric model it requires a large amount of data, which makes it challenging to adopt those methods to other object classes. Due to these limitations, in this paper we propose a hand mesh reconstruction approach which does not utilize a parametric hand model.

\noindent \textbf{Model free 3D hand pose and mesh estimation:}  In the  recent approaches on 3D hand pose and mesh estimation where parametric models are not used, other priors are employed. For an instance using 2D pose of the hand as an input to the network \cite{choi2020pose2mesh}. This requires the annotation of the 2D pose on input images, which limits the approach's ability in adopting to the datasets where the 2D pose annotation is not available. In addition, there exists some approaches that rely on heatmaps of the keypoints at early stages \cite{HAMR,ge20193d}, which requires additional steps of keypoint estimation which can later be extracted directly from the estimated mesh. In contrast, we do not employ 2D or 3D keypoint locations in our method.

\noindent \textbf{Graph Neural Networks (GNNs)  for hand mesh estimation:} In the recent literature, several approaches can be found where GNNs have been employed in estimating the 3D mesh of human hand \cite{choi2020pose2mesh, ge20193d}. However, the objective functions of these methods are limited to the vertex coordinates and other properties associated with the vertex location in the final mesh, where the resultant features of the GCNs are not fully utilized. To fully  harness the strengths of GCNs we incorporate a 3D feature descriptor to our method, where the GCN is aimed to learn not only the vertex locations but it also learn to estimate the 3D feature descriptor, which elevate the overall accuracy of mesh estimation.

\noindent \textbf{Effective use of datasets for hand pose and mesh estimation:}  When the datasets for hand pose and mesh estimation are considered, most recent datasets (Dome dataset by Kulon \textit{et al.} \cite{kulon2019single}, FreiHAND dataset by \textit{et al.} \cite{zimmermann2019freihand}) contains the images and their corresponding groundtruth mesh. These datasets have been used by the state-of-the-art methods for hand pose and mesh estimation \cite{choi2020pose2mesh, moon2020i2l}. However, datasets such as RHD \cite{zb2017hand} and STB \cite{zhang2017hand} contains the images and their corresponding groundtruth 3D pose, and the methods that have used those datasets have targeted only on estimating the 3D pose \cite{ge20193d,zhang2019end, iqbal2018hand, panteleris2018using}. However, we propose an approach where the existing datasets which do not contain the groundtruth mesh details can effectively be used for the task of hand mesh estimation.

\section{Single Image Mesh Generation}

When considering the available datasets for single image hand mesh reconstruction, there are 2 main variations; 1) The datasets which contain images and the corresponding by groundtruth mesh (i.e., the dataset by Kulon \textit{et al.} \cite{kulon2019single} (referred as the Dome dataset hereafter) and the dataset  by  Zimmermann \textit{et al.} \cite{zimmermann2019freihand} (referred as the FreiHAND dataset hereafter), and  2) Other standard datasets such as Rendered Handpose Dataset (RHD) \cite{zb2017hand}  and Stereo Handpose Dataset (STB) \cite{zhang2017hand} that do not contain the groundtruth meshes, instead they contain the 3D and 2D keypoint annotations of the human hand. Therefore we use two different network architectures:  1) To reap the maximum benefit of the availability of the coupled data in the Dome dataset \cite{kulon2019single} and FreiHAND dataset \cite{zimmermann2019freihand} and 2) To use the mesh data in Dome dataset along with the image data in other standard datasets (i.e., STB and RHD) for the robust estimation of the hand mesh.

In this section, we describe the details of our method. First, we introduce the hand mesh representation and the 3D surface feature descriptor we use in this paper. We then elaborate on the methodology that we use for the two aforementioned data settings. 

\subsection{Hand Mesh Representation}

In this work we represent hand mesh $M$ as in Equation \ref{mesh_representation}, where $V$ denotes the vertices  and $F$ denotes the faces that comprises the mesh. Each vertex in $V$ is denoted by its $x$, $y$ and $z$ coordinates (i.e., $v_{i} = \left [ x_{i}, y_{i}, z_{i} \right ]$) and each  face is denoted by the vertex numbers which have contributed for that face (i.e., $f_{i} = \left [v_{p}, v_{q}, v_{r} \right]$).
\begin{equation}
M = \left ( V,F \right ); V\:\epsilon\:\mathbb{R}^{N\times 3}; F\: \epsilon\:V^{M \times 3}
\label{mesh_representation}
\end{equation}

\subsection{3D Surface Descriptor}

In GNNs an attributed graph is defined as,
\begin{equation}
    Graph = \left ( V,E,X \right ),
    \label{graph_representation}
\end{equation}
where $V$ is the set of vertices/nodes which is directly extracted from $M$ (Equation  \ref{mesh_representation}), $E$ is the set of edges which is derived using $F$ in  Equation \ref{mesh_representation}. $X$, can either be node attributes   (i.e., $X^{v} \epsilon\:\mathbb{R}^{N \times d}$ such that $X_{v_{i}} \epsilon\:\mathbb{R}^{d}$, is the feature vector of node $v_{i}$), or edge attributes (i.e., $X^{e} \epsilon\:\mathbb{E}^{T \times c} $ where $T$ is the number of edges in the graph). 

In this work, we use a node feature that can represent distinctive node properties. We selected the Signature of Histogram of Orientations (SHOT) descriptor \cite{salti2014shot}, which has the ability to generate descriptive features for 3D points. SHOT descriptor is a combination of the concepts of ``signature'' \cite{chua1997point} and ``histogram'' \cite{johnson1999using}, such that the descriptor possess computational efficiency while maintaining the robustness. Apart from the evaluations that have been performed by the developers of the SHOT descriptor Samuele \textit{et al.} \cite{salti2014shot}, the SHOT descriptor has demonstrated optimum performance in different domains \cite{guo2016comprehensive}, including in frameworks with deep learning techniques \cite{boscaini2016learning}. The dimension of the feature descriptor depends on the parameters such as the number of neighbours that should be considered at the time of the feature descriptor creation.

%\textcolor{blue}{Ask from Dr. Kien whether I need to write about the construction of the SHOT descriptor}

\subsection{Network Architecture for  Coupled Training Data}
\label{gen_loss_details}
We use a variation of a conditional GAN in this work to generate realistic hand meshes, based on the RGB hand images. The  objective of a conditional GAN is expressed as in Equation \ref{conditional_equation}, where $G$ and $D$ are the functions learned  by the generator and the discriminator respectively.  Conditional GANs are capable of learning the mapping between the input and the desired output. \vspace{-10pt}

\begin{equation} 
\begin{split}
\mathcal{L}_{cGAN}\left ( G,D \right ) & = \mathbb{E}_{x,y}\left [logD\left ( x,y \right )  \right ] \\
 & + \mathbb{E}_{x,z}\left [log\left ( 1- D\left ( x,G\left(x,z \right) \right ) \right ) \right]  
\end{split}
\label{conditional_equation}
\end{equation}

\begin{figure*}[!tb]
\centering
\includegraphics[trim=1 40 1 1,clip, width=\textwidth]{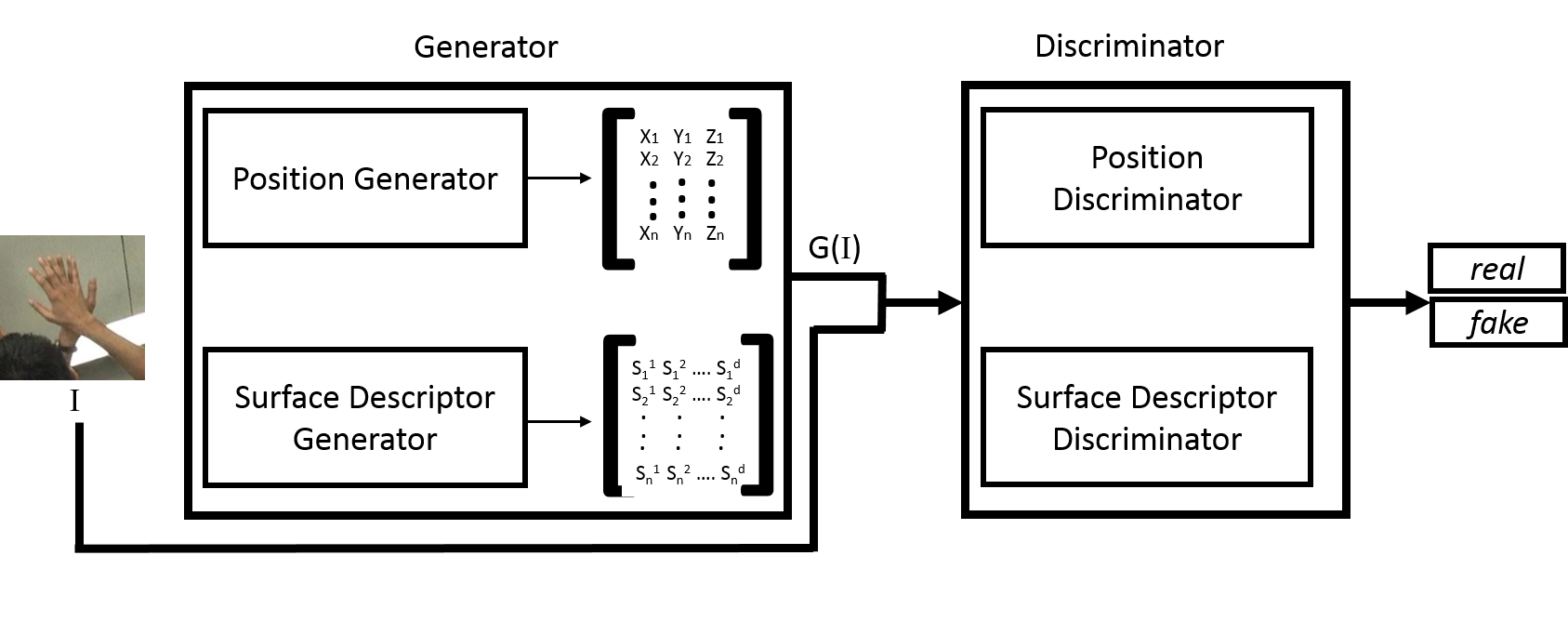}
\caption{The overview of the conditional GAN architecture that is proposed in this paper. The position values and the SHOT descriptor values are generated using the generator network and passed to the discriminator network, which classifies whether they are generated or the groundtruth. \vspace{-20pt} }
\label{config:conditional}
\end{figure*}

The configuration of the conditional GAN that we use is depicted in Figure \ref{config:conditional}. The generator network has 2 components to predict the position vectors (i.e., $V$ in Equation \ref{mesh_representation}) and  the node features (i.e., $X$ in Equation \ref{graph_representation}, wherein this work we have used SHOT descriptor). Similarly, the discriminator network is also composed of 2 components, as ``Position Discriminator'' and ``Surface Descriptor Discriminator''.

 When it comes to  conditional GANs, the generator's objective is to generate output that closely resembles the groundtruth output while fooling the discriminator. Therefore, we define $\mathcal{L}\left ( G \right )$ as in Equation \ref{conditional_gen_loss},

\begin{equation}
\begin{split}
    \mathcal{L}\left ( G \right ) & = \lambda \mathcal{L}_{pos}(G)+ \mu \mathcal{L}_{{shot}}(G) + \theta \mathcal{L}_{normal}(G)\\
    & +\gamma \mathcal{L}_{Laplacian}(G) + \phi \mathcal{L}_{Quadratic}(G),
    \end{split}
    \label{conditional_gen_loss}
\end{equation}
to measure the similarity between the predicted values and the corresponding groundtruth values.

The final objective of the conditional GAN is,
\begin{equation}
G^{*} = arg\;\underset{G}{min}\;\underset{D}{max} \mathcal{L}_{cGAN}\left ( G,D \right )+\delta \mathcal{L}(G) 
\label{gan_objective}
\end{equation}

The first two terms of Equation \ref{conditional_gen_loss} are aimed at minimizing the reconstruction error between  the position vector and the SHOT descriptor respectively. $\mathcal{L}_{pos}$ is defined as, 

\begin{equation}
\mathcal{L}_{pos} = \sum_{i = 1}^{N} \left \| pred_{pos}^{ i}  - gt_{pos}^{i} \right \|_{1},
\label{pos_error}
\end{equation}
where $pred_{pos}$ and $gt_{pos}$ are the predicted and groundtruth vertex locations (i.e., position values) of the mesh \cite{wang2018pixel2mesh}.

We introduce $\mathcal{L}_{shot}$, which is the difference between the surface descriptors (SHOT descriptor) of the groundtruth mesh and the predicted mesh. $\mathcal{L}_{shot}$ is defined as in Equation \ref{shot_error}. 

\begin{equation}
\mathcal{L}_{shot} = \sum_{i = 1}^{N} \left \| pred_{shot}^{ i}  - gt_{shot}^{i}\right \|_{1}
\label{shot_error}
\end{equation}

A loss based on the surface normals of the mesh  is introduced to enforce the smoothness of the mesh. To ensure that the surface normals of the groundruth mesh and the predicted mesh are parallel, the dot product among them is used.  $\mathcal{L}_{normal}$  is calculated as in  Equation \ref{normal_error}, where $n^{i}_{pred}$ and $n^{i}_{gt}$ denote  the normal vector of face $i$ in the predicted mesh and the groundtruth mesh respectively.

\begin{equation}
    \mathcal{L}_{normal} = \sum_{i = 1}^{M} \left \| <n^{i}_{pred} ,n^{i}_{gt}> \right \|_{2}^{2}  
    \label{normal_error}
\end{equation}

In addition, to further enhance the smoothness of the mesh    we employ the Laplacian loss ($\mathcal{L}_{Laplacian}$) \cite{ge20193d}. We introduce two components to the Laplacian loss (Equation \ref{laplacian_components}), where the $\mathcal{L}_{Vertex Laplacian}$ is calculated for each of the vertices in the mesh considering the adjacent neighbours while enforcing the smoothness in a fine grained  context, and $\mathcal{L}_{Keypoint Laplacian}$ is calculated for the keypoints while considering the   neighbours in a broad range, thus enforcing the smoothness in a more coarser level. The weights of the $\mathcal{L}_{Vertex Laplacian}$ and $\mathcal{L}_{Keypoint Laplacian}$ are denoted by $\alpha$ and $\beta$. Laplacian error in general is defined as in Equation \ref{laplacian_errror}, where $w_{i} = pred\_pos_{i} - gt\_pos_{i}$ for $v_{i}$. When considering $\mathcal{L}_{Vertex Laplacian}$, the neighbours of vertex $v_{i}$ are defined as  $\mathcal{N}\left ( v_{i} \right ) = \left \{ w \epsilon V | \left ( v_{i},w \right )\epsilon\:E \right \}$, whereas   $\mathcal{L}_{Keypoint Laplacian}$ is  defined based neighbours that are identified through a graph unrolling and graph traversing process. Please refer to the supplementary material for the details on the method that was used to identify neighbours for   $\mathcal{L}_{Keypoint Laplacian}$ calculation. 
\begin{equation}
    \mathcal{L}_{Laplacian} = \alpha\mathcal{L}_{Vertex Laplacian} + \beta\mathcal{L}_{Keypoint Laplacian}
    \label{laplacian_components}
\end{equation}

\begin{equation}
 Laplacian \; error =  \sum_{i=1}^{N} \left \| \omega_{i} - \sum_{v_{k}\epsilon \:\mathcal{N}\left ( v_{i} \right )} \omega_{k} \middle/ B_{i} \right \| _{2}^{2}  
 \label{laplacian_errror}
\end{equation}

The Quadratic loss ($\mathcal{L}_{Quadratic}$)  (Equation \ref{quadratic loss}) \cite{agarwal2019learning}  is also used to penalize the predicted points in the normal direction. In Equation \ref{quadratic loss}, $Q_{v_{gt}}v_{pred}$ stands for the quadratic error (\cite{garland1997surface,ronfard1996full}) which is calculated based on the triangle incidents that correspond to $v_{gt}$.

\begin{equation}
    \mathcal{L}_{Quadratic} = \frac{1}{N} \sum_{\substack{v_{pred}\epsilon M_{pred} \\ v_{gt}\epsilon M_{gt}}} Q_{v_{gt}}v_{pred}
    \label{quadratic loss}
\end{equation}

In general, the objective of the the discriminator network ($D$ in Equation \ref{conditional_equation}) is to classify whether the given input is from the real sample or whether it has been generated by the generator network ($G$). However, the existing work on GANs is focused on discriminating the generated data such as class  labels and  images and thus have used fully connected or convolutional layers in the discriminator.

In this work we use graph convolutional layers in the ``Surface Descriptor Discriminator'' network (Figure \ref{config:conditional}), where the node features are taken into consideration. As the edge connections ($E$ in Equation \ref{graph_representation}) remain the same for all the estimated meshes we use spectral based graph convolution operations. We used GCN layers introduced by Kim \textit{et al.}  \cite{kipf2016semi}.

\subsection{Network Architecture for Uncoupled Training Data}

\begin{figure*}[!tb]
\centering
\includegraphics[width=\textwidth]{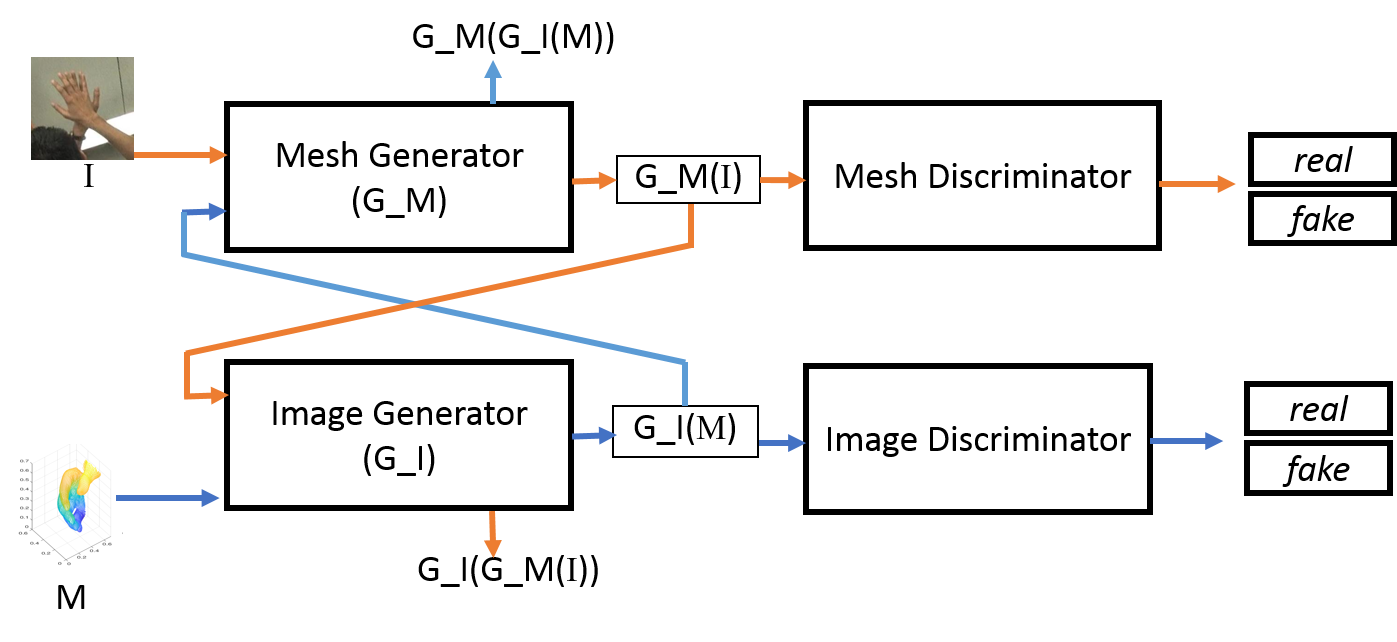}
\caption{The overview of the cycle GAN architecture that is proposed in this paper. $G\_M$ denotes the generator which is designed for estimating the mesh from the input image $I$, where as $G\_I$ denotes the generator which is designed for estimating the image from the input mesh $M$.   }
\label{config:cycle}
\end{figure*}

Hand mesh estimation from a single image suffers from the problem of having a limited amount of  training data and hence the deep learning based techniques with supervised learning can not be used. As a solution many of the existing methods use the datasets with 3D keypoint annotations and estimate the hand pose. The estimated pose is then used along with parametric models such as MANO  \cite{MANO:SIGGRAPHASIA:2017} for the 3D mesh reconstruction.

In this paper, we use a variation of cycle GAN \cite{zhu2017unpaired} to estimate the 3D mesh of hand using a single image, based on uncoupled training data. The overview of the framework that we use in this work is depicted in Figure \ref{config:cycle}, where ``Mesh Generator'' and ``Mesh Discriminator'' consists of position and surface descriptor related components which are denoted in Figure \ref{config:conditional}.

We define the objective function of the network as,
\begin{equation}
\begin{aligned}
    \mathcal{L}\left ( G\_M, G\_I, D\_M, D\_I \right ) &= \mathcal{L}_{GAN}(G\_M, D\_M, I, M)\\ 
    &+ \mathcal{L}_{GAN}(G\_I, D\_I, M, I) \\
    &+ \delta \mathcal{L}_{cyc}(G\_M, G\_I),
    \end{aligned}
    \label{cyclegan_obj}
\end{equation}

\noindent where $\mathcal{L}_{GAN}$ is defined according to the adversarial loss proposed in \cite{goodfellow2014generative} (Equation \ref{original_gan}). $\mathcal{L}_{cyc}$, which stands for cycle consistency loss is used to constraint the possible mapping functions such that the mapping from image to mesh can be made as unique as possible.
We define the cycle consistency loss with two components as $\mathcal{L}_{cyc\_mesh}$ and $\mathcal{L}_{cyc\_im}$, which stand for the cycle consistency of the mesh generator and the image generator respectively. $\mathcal{L}_{cyc\_mesh}$ is defined as in Equation \ref{conditional_gen_loss}, where we aim to retain the surface smoothness of the mesh while minimizing the position error and the shot descriptor error. It should be noted that in this setting $pred$ in Equation \ref{pos_error}, \ref{shot_error}, \ref{normal_error} and \ref{laplacian_components} refers to the $G\_\:M(G\_\:I(M))$ of Figure \ref{config:cycle}.  $\mathcal{L}_{cyc\_im}$ is defined as in Equation \ref{cyc_im}.

%\begin{equation}
%\begin{split}
%    \mathcal{L}_{GAN}\left ( G, D, X,Y  \right ) & = \mathbb{E}_y\left ( logD\left ( y \right )  \\
%    & + \mathbb{E}_x log\left ( 1-D\left ( G\left ( x \right ) \right ) \right )\right )
%\end{split}
%\label{original_gan}
%\end{equation}

\begin{equation}
\begin{split}
L_{GAN}\left ( G,D,X,Y \right ) & = E_{y}\left ( logD\left ( y \right ) \right ) \\ & + E_{x}\left ( log\left ( 1 - D\left ( G\left ( x \right ) \right ) \right ) \right )
\end{split}
\label{original_gan}
\end{equation}

\begin{equation}
    \mathcal{L}_{cyc\_im} =  \mathbb{E}\left \|  G\_\:I(G\_\:M(I))-I\right \|_{1}
    \label{cyc_im}
\end{equation}

\subsection{Generator and Discriminator for Coupled Training Data}
\label{network_coupled}

\noindent The dome dataset \cite{kulon2019single}, contains meshes with 7907 vertices and the FreiHAND dataset, which is based on the MANO model has 778 vertices. The generator network we use is composed with 2 main components 1) To estimate an initial mesh with low resolution and 2) To increase the mesh resolution. For the initial low resolution mesh, we targeted on learning position vectors and SHOT descriptors for 224 vertices. To be compatible with the image shape we derived the shot descriptors with a dimension of 221. For this dataset, SHOT descriptors of a dimension of $221$ were obtained by setting the parameters such that $number\:of\:bins = 7$, $radius\:of\:descriptor\:estimation\:=\:3$  and $minimum\:neighbours\: =\: 3$.\newline

\noindent \textbf{Generator\newline}
For the preliminary layers of the generator we used a convolution based architecture which is in the shape of a U-Net \cite{ronneberger2015u}, however we did not use any skip connections in this work as the domains of the input and output are different. In denoting the 2D convolution layers followed by a batch normalization and a ReLU activation we use the notation of Convolution-BatchNorm-ReLU. We used a similar architecture as in \cite{isola2017image}, where the encoder has 8 layers of Convolution-BatchNorm-ReLUs  with 64, 128, 256, 512, 512, 512, 512 and 512 kernels followed by a decoder with  8 layers of Convolution-BatchNorm-ReLUs with 512,512,512,512,256,128, 64. The above mentioned  convolutions  are $4 \times 4$ filters and decoder network is followed by another convolution layer to make the output channel dimension to 1.  The ReLUs in the encoder are leaky with a slope of 0.2. After the final pass, which results in an output of shape $224 \times 224$ (ignoring the batchsize dimension and channel dimension), we decompose that into two components as $[224,3]$ and $[224,221]$, where the first component is the position vector and the second component is the SHOT descriptor for the $224$ vertices in the coarse grained mesh. \newline\newline
\noindent\textbf{Mesh Enhancer\newline}
\noindent To enhance the mesh resolution, we used a cascade of Multi-branch GCN \cite{qian2019pu} modules, where GCNConv layers \cite{kipf2016semi} were used for feature upsampling.
For the dome dataset we used $5$  Multi-branch GCN  modules at the first cascade level and then $8$ modules in the second cascaded level and for the FreiHAND dataset we use $3$ Multi-branch GCN  modules. The resultant node features which construct mesh at full resolution  were then passed through a set of  Convolution-BatchNorm-ReLU which plays a role analogues to the role of ``Coordinator Reconstructor'' of the initial work on point upsampling \cite{qian2019pu}. This contains Convolution-BatchNorm-ReLU layers with 64, 64, 64, 64 and 1 kernels with each kernel having $1 \times 3$ filters. Thus the output of this network constructs the position vector for the mesh at full resolution.  \newline\newline
\begin{figure*}[!tb]
\centering
\includegraphics[width=\textwidth]{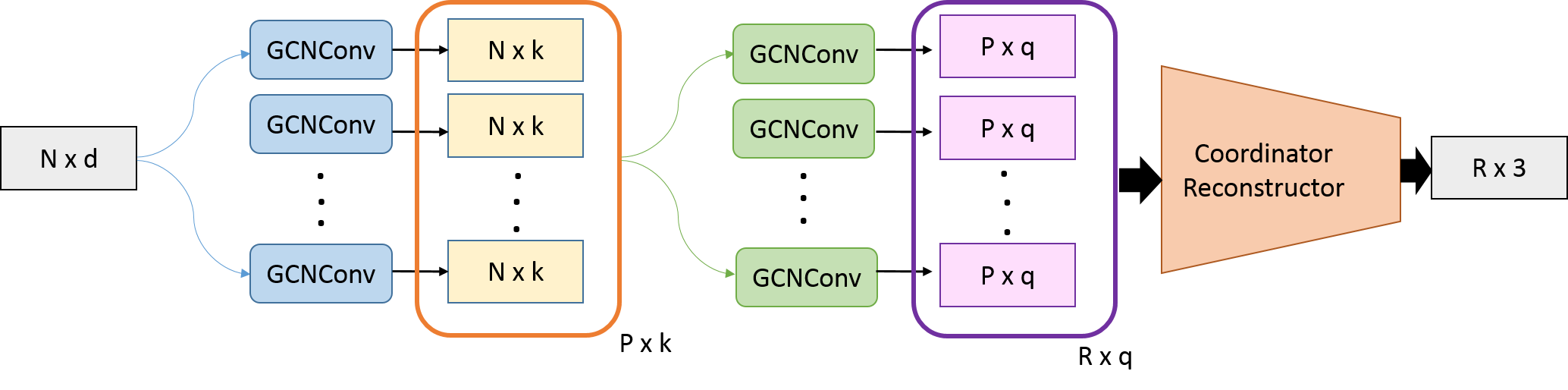}
\caption{The graph enhancement process that was used in our work. It should be noted that this image depicts the process of upsampling a graph which has $N$ nodes and a feature dimension of $d$, to a graph  with $R$ nodes. The depicted network contains two cascaded levels of graph upsampling followed by the ``Coordinate Reconstructor'' which calculates the position vector of the upsampled graph. $k$ and $q$ are the feature dimensions that of the generated features at cascaded level 1 and 2. Since the objective of our work is to upsample the graph while retaining the number of features we set $k = q = d$.\vspace{-10pt}   }
\label{meshenhancer}
\end{figure*}
\noindent\textbf{Discriminator\newline}
\noindent The discriminator network contains 2 branches where the position vectors and the node features are compared with the corresponding groundtruth values. ``Position Discriminator'' takes the input of size $N\times 3$ and the ``Surface Descriptor Discriminator'' takes the input of size $N \times 221$, where $N$ is the number of vertices in the resultant mesh. For the position discriminator we use three Convolution-BatchNorm-ReLU layers with 64,64 and 1 kernels in each with $5 \times 1$ filters and for the ``Surface Descriptor Discriminator'', we use 2 GCNConv layers where we reduce the feature size from $221$ to $100$ and then to $50$. We pass each of these through a fully connected layer with $2048$  nodes and then concatenate the output from the 2 discriminators before passing them trough another fully connected layer of $1048$ nodes followed by a fully connected layer with size 1 and softmax activation.

\subsection{Generator and Discriminator for Uncoupled Training Data}

Compared with the network settings that are described in the above section where the coupled training data is available, when our implementation of cycle GAN (Figure \ref{config:cycle}) is considered the main difference is that in cycle GAN we have enforced the cycle consistency loss where we feed the mesh generated by ``Mesh Generator'' in Figure \ref{config:cycle} to the ``Image Generator'' and vice-versa. To allow this, the inputs for each generator should be same in  shape. Hence we consider the coarse grained mesh which has $224$ vertices and set the input image width and height to $224$ pixels. For the generators and for the ``Image Discriminator'' we used the same architectures that have been used by the original paper of cycle GAN  \cite{zhu2017unpaired}. For the mesh discriminator, we used the same architecture that we described in the ``Discriminator'' subsection of Section \ref{network_coupled}.

For the setting of uncoupled training data, we separately trained the mesh enhancer (Figure \ref{meshenhancer}) such that it learns the mapping from low dimensional mesh to the high dimensional mesh, and the generated  low dimensional meshes were upsampled using it. 

%We used a convolutional network for the estimation of low resolution mesh and then we used a multi branch graph convolution \cite{qian2019pu} to  enhance the resolution of the mesh. For the initial low resolution mesh, we targeted on learning position vectors and SHOT descriptors for 256 vertices. To be compatible with the image shape we derived the shot descriptors with a dimension of $253$. For this dataset, SHOT descriptors of a dimension of $253$ were obtained by setting the parameters such that $number\:of\:bins = 7$, $radius\:of\:descriptor\:estimation\:=\:3$  and $minimum\:neighbours\: =\: 3$.

\subsection{Training}
For the network which uses coupled data we set $\delta = 10$ in Equation \ref{gan_objective}. All the parameters $\lambda$, $\mu$, $\theta$ , $\gamma$ and $phi$ in Equation \ref{conditional_gen_loss}, and  $\alpha$ and $\beta$ in Equation \ref{laplacian_components} were set to 1. To see the effectiveness of the  constraints that were enforced to obtain surface smoothness we evaluate the method with setting the parameters $\theta$ and $\gamma$ to 0. The results related to those settings can be found in the Section \ref{experiment}. For the network which used uncoupled data,  $\delta$ was set to 10. We used a training procedure similar to \cite{isola2017image,zhu2017unpaired} where all the models were trained from the scratch and with a learning rate of 0.0002 using Adam optimizer \cite{kingma2014adam}.

\section{Evaluations}
\label{experiment}
In this section we describe the datasets and the evaluation matrix that we used, the ablative studies that we  conducted to evaluate the effectiveness of the components in our model, the experimental results that we obtained in benchmarking our model with the state-of-the-art methods. It should be noted that when recording the results of the state-of-the-art methods, we have used the evaluations that are been performed by the respective authors and the results with the best configuration of their proposed methods have been considered. 

\subsection{Datasets}
This work utilizes two types of datasets, 1) Coupled dataset where the images and the corresponding groundtruth mesh is available and 2) Datasets which contain only the images. For the coupled dataset we used the Dome dataset \cite{kulon2019single} and FreiHAND dataset \cite{zimmermann2019freihand} and for the later we used the RHD \cite{zb2017hand}   and STB \cite{zhang2017hand} hand datasets with the 3D meshes which are available as the groundtruth in Dome dataset.

\subsection{Evaluation Protocol}
For the Dome dataset we used the L1 reconstruction error,  between the ground truth mesh and the predicted mesh. For the quantitative evaluations of the   datasets for which groundtruth meshes are not available we extracted the 3D locations of the keypoints and used the accuracy measurement of Percentage of Correct Keypoints (PCK) scores. In PCK calculation, if the predicted keypoint, which we extract from the estimated mesh lies within a sphere with a specific radius with respect to the groundtruth value, it is considered as correct keypoint. 

\subsection{Ablative Studies}

Our ablative studies were conducted on the Dome dataset which has the groundtruth mesh data, as such the quantitative evaluations could be carried out. We conducted the ablative studies, with the aim of evaluating the contribution of the components in Equation \ref{conditional_gen_loss}. All the results related to  ablative studies are recorded in Table \ref{ablative}.  \newline

\noindent \textbf{Effectiveness of using the 3D surface descriptor:}
We performed this ablative study to evaluate the effectiveness of incorporating  $\mathcal{L}_{shot}$, which measures the similarity in the groundtruth and generated SHOT descriptor. 
We trained our model on training subset and tested on the test subset of the Dome dataset \cite{kulon2019single}. For this ablative study, we set  $\mu = 0$ in Equation \ref{conditional_gen_loss}.\newline 

\begin{figure}[!tb]
\centering

\begin{tabular}{cc}
  \includegraphics[width=40mm]{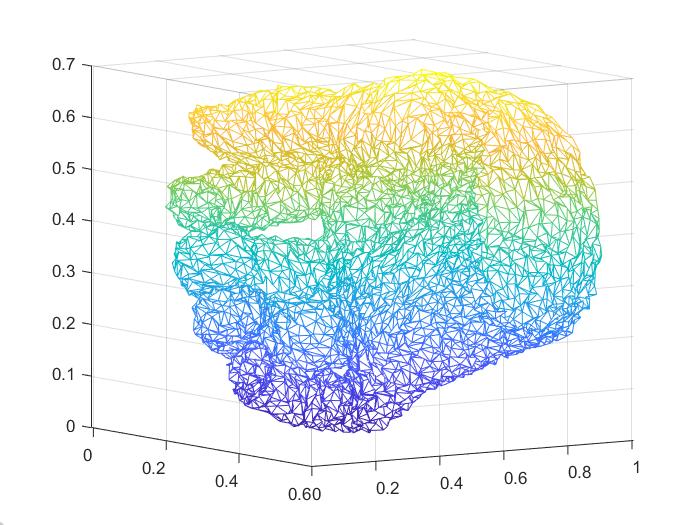} &   \includegraphics[width=40mm]{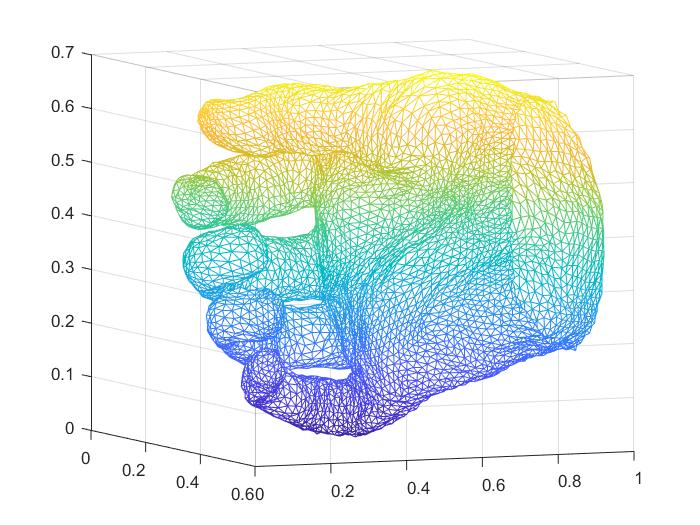} \\
(a) $\theta =0$ and $\gamma=0$ & (b) $\theta =0$ and $\gamma=1$ \\[6pt]
 \includegraphics[width=40mm]{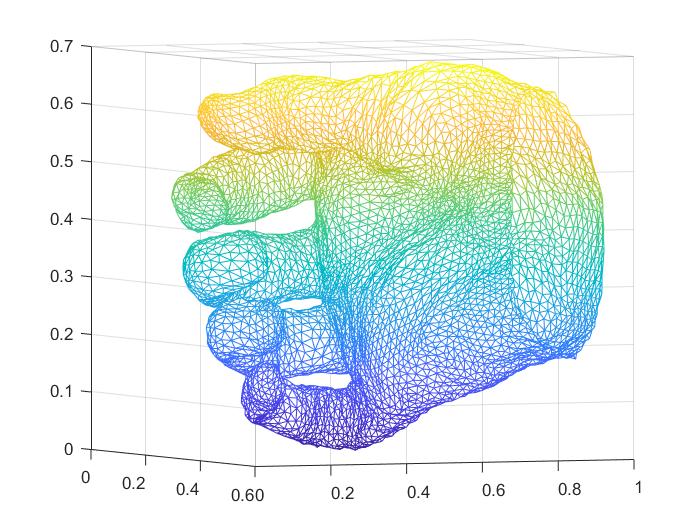} &   \includegraphics[width=40mm]{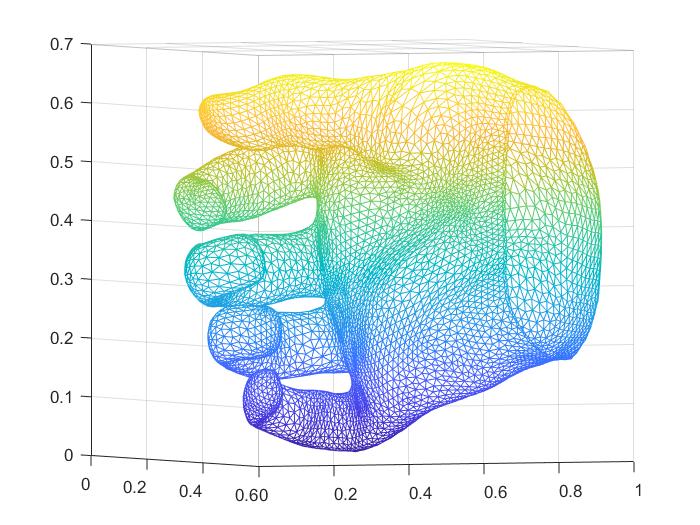} \\
(c) $\theta =1$ and $\gamma=0$ & (d) $\theta =1$ and $\gamma=1$ 
\end{tabular}
\caption{The qualitative results that were obtained by changing the parameters that are related to surface smoothness, in Equation \ref{conditional_gen_loss}. \vspace{-20pt}}
\label{smoothness_ablative}
\end{figure}

\noindent \textbf{Effectiveness of enforcing the surface smoothness in the mesh:}
As described in Section \ref{gen_loss_details}, our method combines several loss functions to enforce the surface smoothness of the mesh. We evaluate the effectiveness of each of these loss components (i.e., $\mathcal{L}_{normal}$ and  $\mathcal{L}_{Laplacian}$, where the later  consists of 2 components as $\mathcal{L}_{VertexLaplacian}$ and $\mathcal{L}_{KeypointLaplacian}$). We assessed the effectiveness of using components individually and in combination. First we eliminate all the loss values that are related to the surface smoothness, and the corresponding results can be found in the second raw of Table \ref{ablative}. Similarly by eliminating individual lossless we compared the reconstruction error (Table \ref{ablative}). Figure \ref{smoothness_ablative} better visualizes the effect of constraints that are used to smooth the surface. From the conducted ablative studies it is evident that the contribution of the components in the loss function are vital and the compound of all the components has resulted in better accuracy values. 

\begin{table}[!b]
\begin{center}

%\begin{tabular}{lll}
\begin{tabular}{|p{3cm}|p{2cm}|c|}
\hline
Objective & Parameter Values & \makecell{Reconstruction\\ Error \\ (mm)}\\ \hline \hline
Effectiveness of using the 3D surface descriptor &$\lambda = 1$, $\mu = 0$, $\theta=1$, $\gamma=1$, $\alpha=1$, $\beta=1$, $\phi=1$	& 2.64\\
Effectiveness of enforcing the surface smoothness	& $\lambda = 1$, $\mu = 1$, $\theta=0$, $\gamma=0$, $\phi=1$, $\alpha$ and $\beta$ not applicable	& 2.96\\
Effectiveness of surface normal error & $\lambda = 1$, $\mu = 1$, $\theta=0$, $\gamma=1$, $\alpha=1$, $\beta=1$, $\phi=1$	& 2.88\\
Effectiveness of Laplacian error	&$\lambda = 1$, $\mu = 1$, $\theta=1$, $\gamma=0$, $\phi=1$, $\alpha$ and $\beta$ not applicable		& 2.76\\
Effectiveness of vertex Laplacian error&	$\lambda = 1$, $\mu = 1$, $\theta=1$, $\gamma=1$, $\alpha=0$, $\beta=1$, $\phi=1$ &	2.73\\
Effectiveness of keypoint Laplacian error	& $\lambda = 1$, $\mu = 1$, $\theta=1$, $\gamma=1$, $\alpha=1$, $\beta=0$,  $\phi=1$ &	2.69\\
Effectiveness of the total loss	& $\lambda = 1$, $\mu = 1$, $\theta=1$, $\gamma=1$, $\alpha=1$, $\beta=1$, $\phi=0$ &	\textbf{1.83}\\
Effectiveness of the total loss	& $\lambda = 1$, $\mu = 1$, $\theta=1$, $\gamma=1$, $\alpha=1$, $\beta=1$, $\phi=1$ &	\textbf{1.83}\\
\hline
\end{tabular}
\end{center}
\caption{Results of the conducted ablative studies to evaluate the effectiveness of the loss components that are indicated in Equation \ref{conditional_gen_loss} and in Equation \ref{laplacian_components}.}
\label{ablative}
\end{table}

\subsection{Comparison to the State-of-the-art}

For the Dome dataset we compared our method with the state-of-the-art method which has used the same dataset. When the two methods were compared using the L1 reconstruction error, the obtained results are recorded in Table \ref{dome_stoa}. From the obtained values it can be seen that our method has outperformed the state-of-the-art method with significant margins. Similarly for FreiHAND dataset, we compared our results with the state-of-the-art methods and the obtained results are recorded in Table \ref{freihand_stoa}
Please refer to supplementary material where we have provided  qualitative results on the Dome dataset and FreiHAND dataset.  

\begin{figure*}[!tbh]
\centering
  \begin{subfigure}[b]{0.35\textwidth}
    \includegraphics[width=\textwidth]{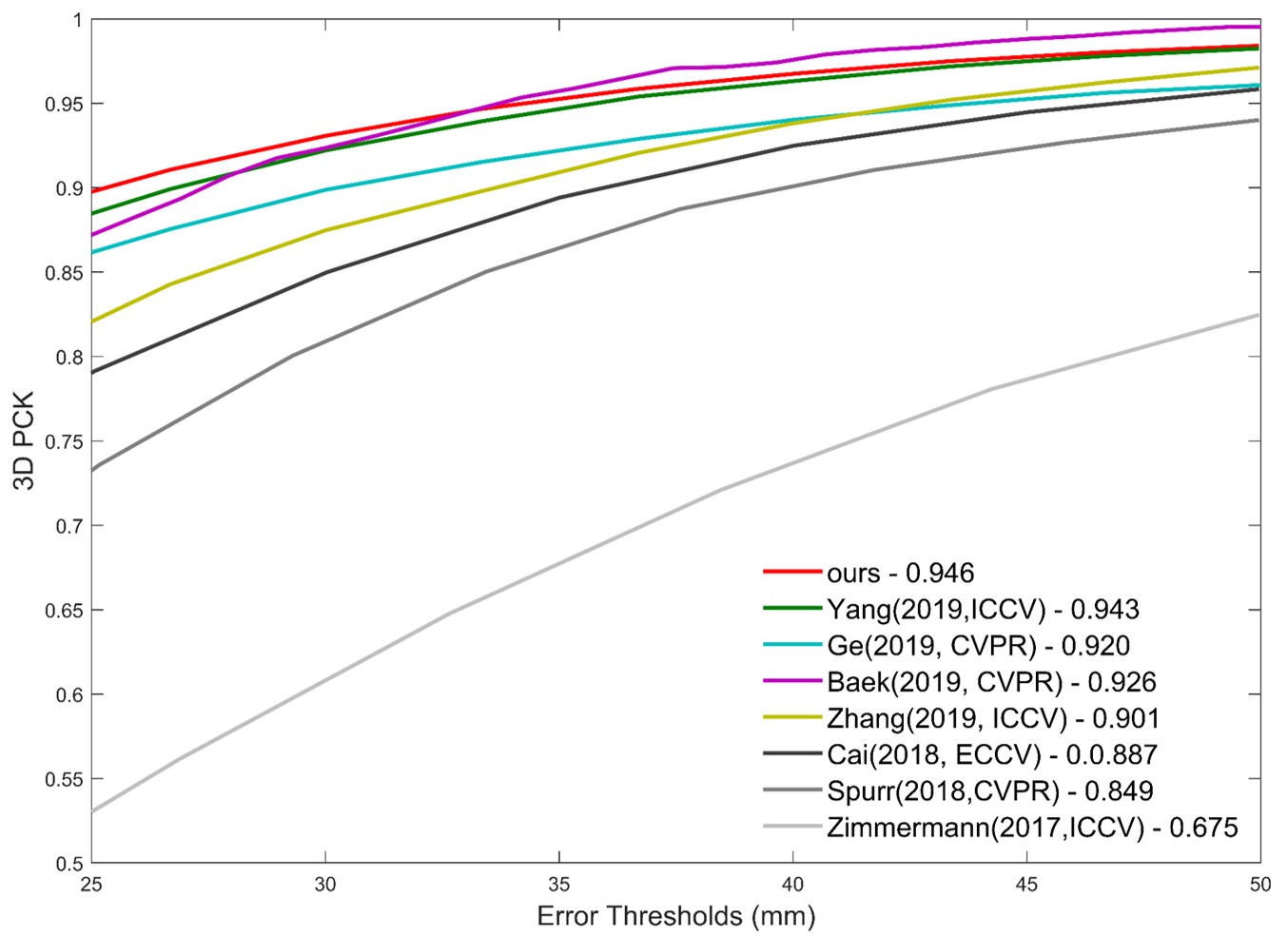}
    \caption{3D PCK values on RHD dataset .}
    \label{rhd_stoa}
  \end{subfigure}
  \begin{subfigure}[b]{0.35\textwidth}
    \includegraphics[width=\textwidth]{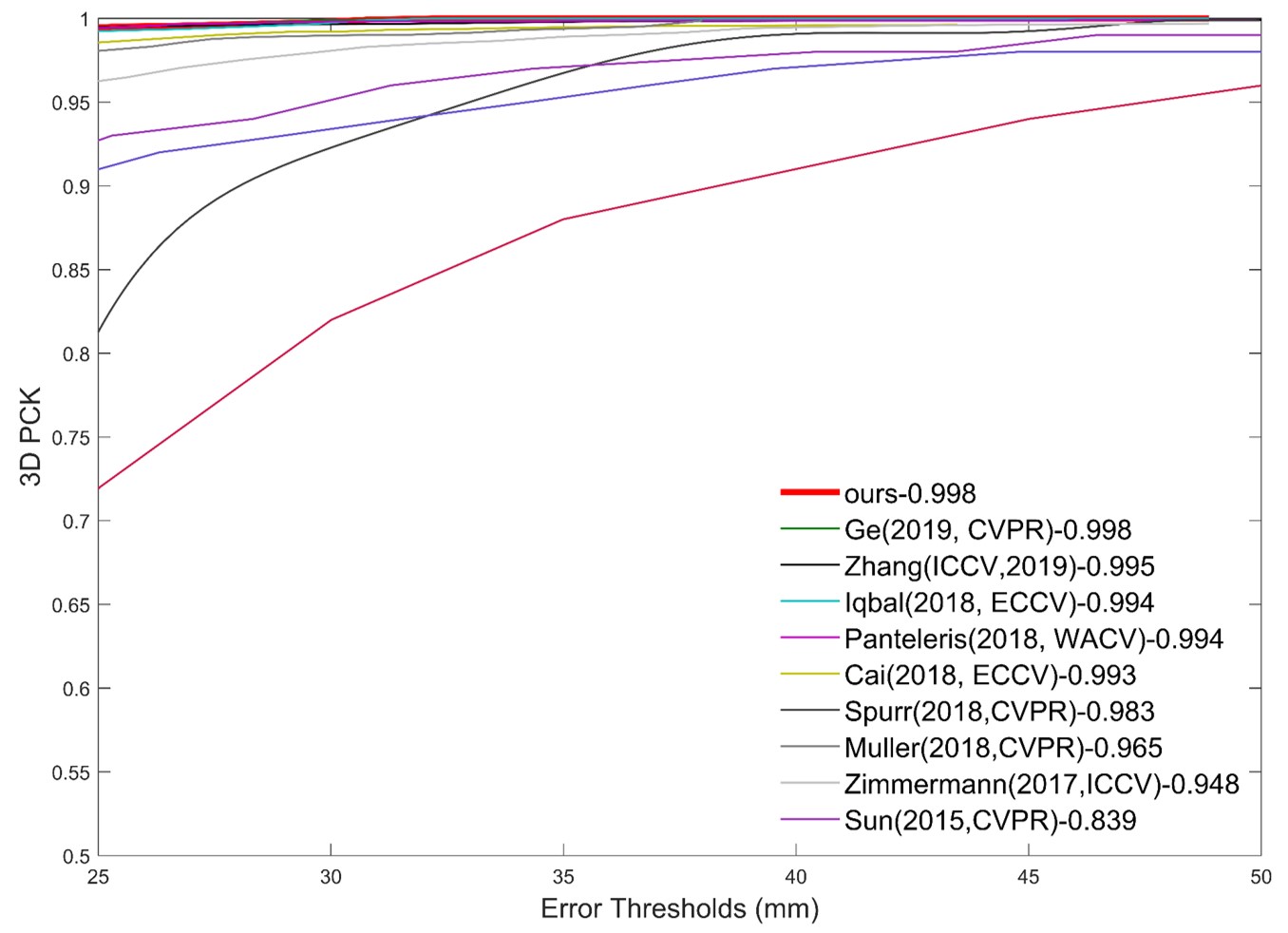}
    \caption{3D PCK values on STB dataset.}
    \label{stb_stoa}
  \end{subfigure}
  \centering
  \caption{The PCK values that were obtained when comparing our method with the state-of-the-art methods, where our method has comparable results to the state-of-the-art methods.\vspace{-10pt}}
  \label{pck_values}
\end{figure*}

For the RHD dataset and STB dataset, the evaluations were performed using the 3D PCK values, where we compare the extracted 3D keypoint values with the groundtruth 3D keypoint values. The obtained PCK values are recorded in Figure \ref{pck_values}. From the results it can be seen that, though the objective of our method was not estimating the 3D keypoint locations, with our method we have obtained PCK values which are comparable to the state-of-the-art methods. It should be noted that our method has not been trained with 3D pose supervision. However, it outperforms the state-of-the-art methods that have been trained with 3D pose supervision. Please refer to the supplementary material for the Area Under Curve (AUC) values related to Figure \ref{rhd_stoa}  and  Figure \ref{stb_stoa}.

Additionally, the in supplementary material we have provided qualitative results for   the 3D reconstructions that were obtained by our method in the setting of uncoupled training data. We have illustrated the  low resolution meshes as well as the high resolution meshes which were generated by our method.

\begin{table}
\begin{center}

%\begin{tabular}{lll}
\begin{tabular}{|p{5cm}|c|}
\hline
Method &  \makecell{Reconstruction\\ Error (mm)}\\ \hline \hline
Kulon \textit{et al.}  (BMVC, 2019) \cite{kulon2019single}& 2.33\\
Ours	& \textbf{1.83}\\
\hline
\end{tabular}
\end{center}
\caption{L1 Reconstruction error which was obtained for the Dome dataset \cite{kulon2019single}.}
\label{dome_stoa}
\end{table}

\begin{table}[]
\begin{center}

\begin{tabular}{|l|c|c|c|}
\hline
Method                & PA MPVPE & F@5mm & F@15mm \\ \hline
\textit{Hasson et al} \cite{hasson2019learning} & 13.2     & 0.436 & 0.908  \\
Boukhayma \textit{et al.} \cite{3DHandShapePoseWild}     & 13       & 0.435 & 0.898  \\
FreiHAND \cite{zimmermann2019freihand}          & 10.7     & 0.529 & 0.935  \\
Pose2Mesh    \cite{choi2020pose2mesh}         & 7.8      & 0.674 & 0.969  \\
I2L-MeshNet   \cite{moon2020i2l}        & 7.6      & 0.681 & 0.973  \\
Im2Mesh GAN (Ours)    &          &       &        \\ \hline
\end{tabular}
\end{center}
\caption{The PA MPVPE and F-scores which were obtained for the FreiHAND dataset \cite{zimmermann2019freihand}.}
\label{freihand_stoa}
\end{table}

Additionally, in supplementary material we have  provided qualitative results for the 3D reconstructions that were obtained by our method in the setting of uncoupled training data. We have illustrated the  low resolution meshes as well as the high resolution meshes which were generated by our method.

\section{Conclusion}

While 3D mesh reconstruction of the human hand using a single image has been explored in the past, the problem still remains a challenge due to the high degree of freedom of the human hand. In this paper, we have presented a method to create 3D mesh of the hand using a single image that can effectively use the existing databases for better reconstruction of the 3D mesh using a single image. We have designed a loss function that can generate more realistic hand meshes, and we demonstrate the effectiveness of that loss function in two settings of Generative Adversarial Networks. The first setting is targeted on the effective use of coupled datasets where the groundtruth meshes are available, whereas the second setting is targeted on uncoupled datasets. In addition, we employ a 3D surface descriptor in this work along with graph convolution networks, which enable the preservation of the surface details of generated meshes. We confirm that our framework outperforms the state-of-the-art as well as the first effort to incorporate explicit 3D features in a single image-based 3D mesh reconstruction.  One of the interesting properties of the proposed mesh recovery approach is that there is no need for parametric hand models as priors. The geometry of the hand is learned and encoded directly in the generator through the end-to-end adversarial training process. This fact enables the proposed algorithm to be easily adapted to other mesh problems such as other body parts or 3D objects.

\iffalse
\begin{figure*}[!b]
\centering
\begin{tabular}{cccc}
  \includegraphics[width=20mm]{rhd-im.png} &   \includegraphics[width=20mm]{rhd-1-3.png} &
  \includegraphics[width=20mm]{rhd-1-1.png} &
  \includegraphics[width=20mm]{rhd-1-2.png}\\
  \includegraphics[width=20mm]{rhd-2-im.png} &   \includegraphics[width=20mm]{rhd-2-3.png} &
  \includegraphics[width=20mm]{rhd-2-1.png} &
  \includegraphics[width=20mm]{rhd-2-2.png}\\
(a) Input Image &  \makecell{ (b) Low Resolution\\ Mesh} &  \makecell{(c) Ours \\ (Perspective 1)} &  \makecell{(s) Ours \\ (Perspective 2)}
\end{tabular}
\caption{The qualitative results that were obtained for the model with uncoupled training data.}
\label{rhd_qualitative}
\end{figure*}

\fi

{\small
\bibliographystyle{ieee_fullname}
\bibliography{egbib}
}

\end{document}